\documentclass{article}
\usepackage{authblk}
\usepackage[utf8]{inputenc}
\usepackage{main}
\usepackage{microtype}
\usepackage{subcaption}
\usepackage{graphicx}
\usepackage{times}
\usepackage{latexsym}
\usepackage{amsmath}
\usepackage{float}
\usepackage{footnote}
\usepackage{enumitem}
\usepackage{bm}
\usepackage{arydshln}
\usepackage{booktabs}
\usepackage{multicol}
\usepackage{multirow}
\usepackage{color}
\usepackage{xcolor}     
\usepackage{colortbl}
\usepackage{bbding}
\usepackage{makecell}
\usepackage{mathtools}
\usepackage{imakeidx}
\usepackage{longtable}
\usepackage{wrapfig}
\usepackage{rotating}
\makeindex
\usepackage{arydshln}
\usepackage{lipsum}
\usepackage{natbib}
\usepackage[toc]{multitoc}
\usepackage[edges]{forest}
\usepackage[normalem]{ulem}
\definecolor{mydarkblue}{rgb}{0,0.08,0.45}
\usepackage[colorlinks=true,linkcolor=mydarkblue,citecolor=mydarkblue,filecolor=mydarkblue,urlcolor=mydarkblue]{hyperref}
\usepackage{CJKutf8}
\usepackage{awesomebox} 
\usepackage{bbding}
\usepackage[most]{tcolorbox}
\usepackage{booktabs}
\usepackage{geometry}
\definecolor{wkblue}{rgb}{0.2, 0.3, 0.6}
\definecolor{meta-color}{rgb}{0.5, 0.5, 0.5}
\usepackage{amsmath}
\usepackage{enumitem}
\usepackage{lscape} 
\usepackage{booktabs}
\usepackage{algorithm}  
\usepackage{pifont}

\usepackage{algorithmicx}
\usepackage{algpseudocode}
\usepackage{tabularx,booktabs}
\usepackage{makecell}

\usepackage{amssymb}
\usepackage{amsfonts}

\usepackage[tikz]{bclogo}
\usepackage[framemethod=tikz]{mdframed}
\definecolor{bgblue}{RGB}{245,243,253}
\definecolor{ttblue}{RGB}{91,194,224}
\definecolor{codegreen}{rgb}{0,0.6,0}
\definecolor{codegray}{rgb}{0.5,0.5,0.5}
\definecolor{codepurple}{rgb}{0.58,0,0.82}
\definecolor{backcolour}{rgb}{0.95,0.95,0.92}
\definecolor{wkgreen}{RGB}{184,244,175}
\definecolor{wkpurple}{RGB}{210,210,253}
\definecolor{wkyellow}{RGB}{255,241,177}
\definecolor{wkblue}{RGB}{174,217,253}
\definecolor{mydarkblue}{rgb}{0,0.08,0.45}
\definecolor{wkblue}{rgb}{0.2, 0.3, 0.6}
\definecolor{dpblue}{rgb}{0.0, 0.333, 0.643}

\mdfdefinestyle{mystyle}{%
  rightline=true,
  innerleftmargin=10,
  innerrightmargin=10,
  outerlinewidth=3pt,
  topline=false,
  rightline=true,
  bottomline=false,
  skipabove=\topsep,
  skipbelow=\topsep
}

\newtcolorbox{myboxi}[1][]{
  breakable,
  title=#1,
  colback=red!5,
  colbacktitle=red!5,
  coltitle=black,
  fonttitle=\bfseries,
  bottomrule=0pt,
  toprule=0pt,
  leftrule=2pt,
  rightrule=2pt,
  titlerule=0pt,
  arc=0pt,
  outer arc=0pt,
  colframe=red,
}

\newtcolorbox{myboxnote}[1][]{
  breakable,
  title=#1,
  colback=orange!0,
  colbacktitle=orange!0,
  coltitle=black,
  fonttitle=\bfseries,
  bottomrule=0pt,
  toprule=0pt,
  leftrule=2pt,
  rightrule=2pt,
  titlerule=0pt,
  arc=0pt,
  outer arc=0pt,
  colframe=orange,
}

\newtcolorbox{myboxii}[1][]{
  breakable,
  freelance,
  title=#1,
  colback=white,
  colbacktitle=white,
  coltitle=black,
  fonttitle=\bfseries,
  bottomrule=0pt,
  boxrule=0pt,
  colframe=white,
  overlay unbroken and first={
  \draw[red!75!black,line width=3pt]
    ([xshift=5pt]frame.north west) -- 
    (frame.north west) -- 
    (frame.south west);
  \draw[red!75!black,line width=3pt]
    ([xshift=-5pt]frame.north east) -- 
    (frame.north east) -- 
    (frame.south east);
  },
  overlay unbroken app={
  \draw[red!75!black,line width=3pt,line cap=rect]
    (frame.south west) -- 
    ([xshift=5pt]frame.south west);
  \draw[red!75!black,line width=3pt,line cap=rect]
    (frame.south east) -- 
    ([xshift=-5pt]frame.south east);
  },
  overlay middle and last={
  \draw[red!75!black,line width=3pt]
    (frame.north west) -- 
    (frame.south west);
  \draw[red!75!black,line width=3pt]
    (frame.north east) -- 
    (frame.south east);
  },
  overlay last app={
  \draw[red!75!black,line width=3pt,line cap=rect]
    (frame.south west) --
    ([xshift=5pt]frame.south west);
  \draw[red!75!black,line width=3pt,line cap=rect]
    (frame.south east) --
    ([xshift=-5pt]frame.south east);
  },
}

\usepackage{fancyhdr} 
\usepackage{blindtext} 
\usepackage{makecell}

\newcommand{\modelname}{\textsc{ToRL}\xspace}

\DeclareCaptionFont{black}{\color{black}}

\definecolor{myblue}{rgb}{0.9, 0.1, 0.94}
\definecolor{mygreen}{rgb}{0.64, 0.56, 0.88}
\definecolor{myyellow}{rgb}{0.68, 0.6, 0.1}
\definecolor{fancygreen}{rgb}{0.33, 0.68, 0.20}
\definecolor{salmon}{rgb}{0.94, 0.52, 0.49}
\definecolor{tablegreen}{rgb}{0.82, 0.94, 0.75}
\definecolor{tableblue}{rgb}{0.81, 0.90, 0.94}
\definecolor{tablered}{rgb}{0.97, 0.85, 0.85}
\definecolor{tableorange}{rgb}{0.96, 0.85, 0.81}
\definecolor{tir}{rgb}{0.592,0.741,0.988}
\definecolor{cot}{rgb}{0.965,0.443,0.537}
\definecolor{grey}{rgb}{0.502,0.502,0.502}
\definecolor{grey}{rgb}{0.502,0.502,0.502}
\definecolor{darkyellow}{rgb}{0.855,0.647,0.125}

\newenvironment{itemize*}%
 {\leftmargini=10pt\begin{itemize}%
  \setlength{\itemsep}{0pt}%
  \setlength{\parskip}{0pt}%
  }%
 {\end{itemize}}
\newenvironment{enumerate*}%
 {\begin{enumerate}%
  \setlength{\itemsep}{0pt}%
  \setlength{\parskip}{0pt}}%
 {\end{enumerate}}

\usepackage{listings}

\newcommand\JSONnumbervaluestyle{\color{blue}}
\newcommand\JSONstringvaluestyle{\color{red}}

\newif\ifcolonfoundonthisline

\makeatletter

\lstdefinestyle{json}
{
  showstringspaces    = false,
  keywords            = {false,true},
  alsoletter          = 0123456789.,
  morestring          = [s]{"}{"},
  stringstyle         = \ifcolonfoundonthisline\JSONstringvaluestyle\fi,
  MoreSelectCharTable =%
    \lst@DefSaveDef{`:}\colon@json{\processColon@json},
  basicstyle          = \ttfamily,
  keywordstyle        = \ttfamily\bfseries,
}

\newcommand\processColon@json{%
  \colon@json%
  \ifnum\lst@mode=\lst@Pmode%
    \global\colonfoundonthislinetrue%
  \fi
}

\lst@AddToHook{Output}{%
  \ifcolonfoundonthisline%
    \ifnum\lst@mode=\lst@Pmode%
      \def\lst@thestyle{\JSONnumbervaluestyle}%
    \fi
  \fi
  \lsthk@DetectKeywords%
}

\lst@AddToHook{EOL}%
  {\global\colonfoundonthislinefalse}

\makeatother

\usepackage{etoolbox}
\usepackage{natbib}
\usepackage{url}
\newcounter{bibcount}
\makeatletter
\patchcmd{\@lbibitem}{\item[}{\item[\hfil\stepcounter{bibcount}{[\thebibcount]}}{}{}
\renewcommand\NAT@bibsetup%
  [1]{\setlength{\leftmargin}{\bibhang}\setlength{\itemindent}{-\parindent}%
      \setlength{\itemsep}{\bibsep}\setlength{\parsep}{\z@}}
\makeatother

\newcommand*\samethanks[1][\value{footnote}]{\footnotemark[#1]}

\definecolor{mybrown}{RGB}{128,64,0}

\definecolor{titlecolor}{HTML}{4c9cff}

\begin{document}



\title{\modelname: Scaling Tool-Integrated RL}

\author{%
\textbf{Xuefeng Li\thanks{~~Co-first authors}\space\space\space 
Haoyang Zou\samethanks\hspace{0.5em}
Pengfei Liu}\thanks{~~Corresponding author}\\
SJTU, SII, GAIR}
  
\maketitle
\thispagestyle{fancy}
\fancyhead{}
\lhead{\includegraphics[height=0.67cm]{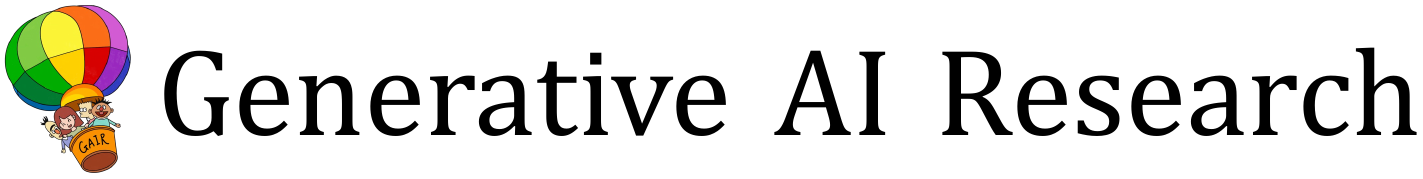}}
\renewcommand{\headrulewidth}{0pt}
\setlength{\headsep}{0mm}

\begin{abstract}

We introduce \textbf{\modelname} (Tool-Integrated Reinforcement Learning), a framework designed to train large language models (LLMs) to autonomously utilize computational tools through \textbf{scaling reinforcement learning directly from \emph{base} models} (i.e., models without post-training). Unlike approaches based on supervised fine-tuning, \modelname enables models to discover optimal strategies for tool utilization via unrestricted exploration. Experiments with Qwen2.5-Math base models demonstrate substantial improvements, with \modelname-7B achieving \textbf{43.3\%} accuracy on AIME24—surpassing RL without tool integration by \textbf{14\%} and the best existing Tool-Integrated Reasoning (TIR) model by \textbf{17\%}. Further analysis reveals several emergent \emph{cognitive behaviors}—such as strategic tool invocation, self-regulation of ineffective code generation, and dynamic adaptation between computational and analytical reasoning. These capabilities \textbf{emerge without explicit instruction}, solely through reward-driven learning. We open-source our implementation, datasets, and models at \url{https://github.com/GAIR-NLP/ToRL}.

\end{abstract}
\begin{figure}[htbp]
    \centering
    
    \begin{subfigure}{\textwidth}
        \centering
        \includegraphics[width=0.9\textwidth]{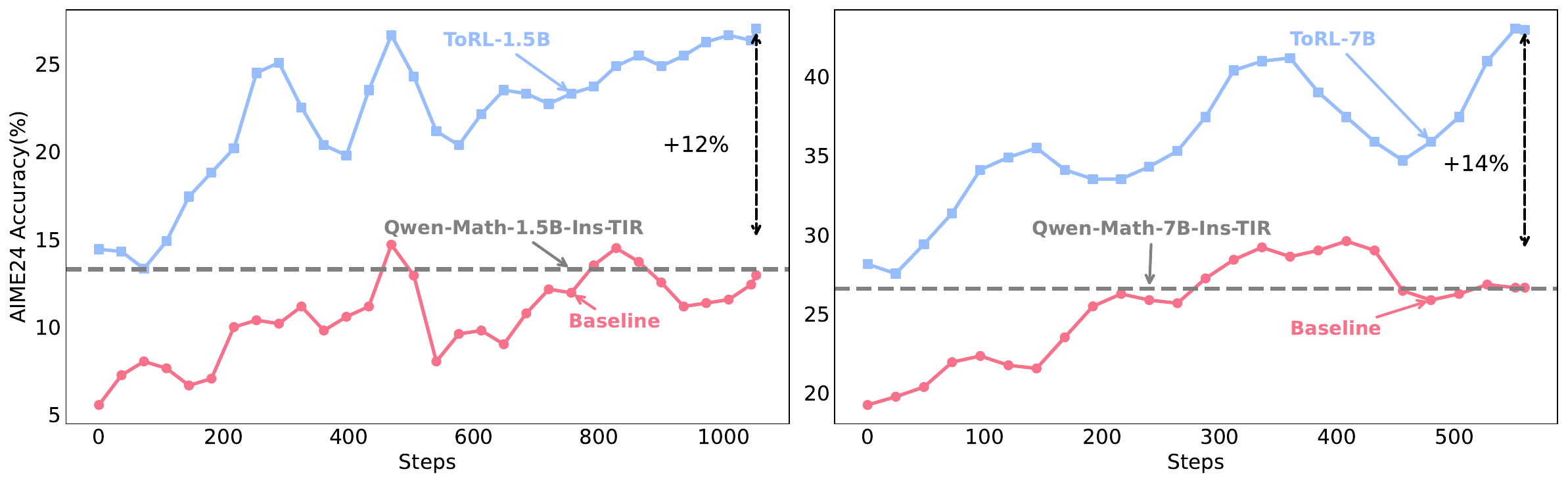} 
        \label{subfig:performance}
    \end{subfigure}
    
    \vspace{1em} 
    
    \begin{subfigure}{0.90\textwidth}
        \centering
        \fontsize{7}{9}\selectfont 
        \begin{tabular}{l}
        \toprule 
        Question: Two parabolas are the graphs of the equations  $y=2x^2$ \ldots Please give the value of k + m. \\
        \midrule
        Response: To solve this problem, we need to find the points of intersection between  \ldots \\
        \texttt{```python} \\
        \texttt{k\_m\_sum = k + m; \textcolor{darkyellow}{print}(k\_m\_sum)} \\
        \texttt{```} \\
        \texttt{```output} \textbf{\texttt{16}} \texttt{```} \\
        \ldots \textcolor{red}{\(k + m = -1 + (-8) = -9\). \ldots} \textcolor{red}{However, the output of the code is 16, \ldots Let's recheck the code and the steps:} \\
        \texttt{```python} \\
        \ldots \\
        \textcolor{fancygreen}{\texttt{\# Find the corresponding y-coordinates}} \\
        \texttt{intersection\_points = [(x\_val, eq1.\textcolor{darkyellow}{subs}(x, x\_val)) \textcolor{ttblue}{for} x\_val \textcolor{ttblue}{in} intersection\_x]}\ldots \\
        \texttt{\textcolor{darkyellow}{print}((intersection\_points, k\_m\_sum))} \\
        \texttt{```} \\
        \texttt{```output} \textbf{\texttt{([(-2, 18), (8, 38)], 16)}}  \texttt{```} \\
        Thus, the value of \(k + m\) is \(\boxed{16}\). \\
        \bottomrule
        
        \end{tabular}
        \label{subfig:behavior}
    \end{subfigure}
    
    \caption{Top Figure: Performance comparison of \modelname versus baseline models(16-step moving). Both plots show AIME24 Accuracy (\%) against training steps across 1.5B and 7B models. In both cases, \textcolor{tir}{\modelname(Ours)} significantly outperforms the \textcolor{cot}{baseline without tool} and \textcolor{grey}{Qwen-2.5-Math-Instruct-TIR}, achieving up to 12\%(1.5B) and 14\%(7B) higher. Bottom Figure: Emergent cognitive behavior during training. \modelname first cross-validates the tool's output with reasoning results. Upon detecting inconsistencies, it engages in reflection and further verification through tool.}
    \label{fig:combined}
\end{figure}

\newpage

\pagestyle{fancy}
\lhead{\rightmark}
\renewcommand{\headrulewidth}{0.7pt}
\setlength{\headsep}{5mm}

\clearpage

\newpage

\section{Introduction}


Recent advances in large language models (LLMs) have demonstrated remarkable reasoning capabilities through reinforcement learning (RL). Models such as o1, R1 and Kimi exhibit emergent abilities including backtracking, self-correction, and reflection when trained with large-scale RL ~\citep{Openo1, guo2025deepseek, kimiteam2025kimik15scalingreinforcement}. Several open-source efforts have shown similar promising results~\cite{qin2024o1, huang2024o1,zeng2025simplerl, deepscaler2025,ye2025limo}.

Traditional reasoning in language models has relied on pure natural language approaches such as Chain-of-Thought (CoT)~\cite{wei2022chain}. While effective for many tasks, these methods falter when facing complex calculations, equation solving, or processes requiring precise computation~\cite{gao2023pal, chen2022program}. Despite research into code generation models, few approaches successfully bridge the gap between LLMs' reasoning capabilities and computational tools' execution power through reinforcement learning.
Tool-Integrated Reasoning (TIR), explored prior to recent RL scaling breakthroughs, enables models to invoke external tools by writing code, executing it through interpreters, and iteratively generating reasoning informed by code outputs. Implementations like ToRA, MathCoder, and Qwen2.5-Math-Instruct-TIR demonstrate TIR's effectiveness in enhancing mathematical problem-solving~\cite{gou2023tora, wang2023mathcoder, shao2024deepseekmath, liao2024mario, yang2024qwen2}.

Despite these advances, existing TIR approaches face critical limitations. Most studies distill trajectories from stronger models and perform Supervised Fine-Tuning (SFT), restricting models to predetermined tool usage patterns and limiting exploration of optimal strategies. While some work, like Qwen-Math, applies RL to SFT-trained models, limited implementation transparency obscures understanding of tool integration within RL frameworks. 
To address these challenges, we introduce \modelname (Tool-Integrated Reinforcement Learning), a framework that scales reinforcement learning directly from base models without the constraints of prior supervised fine-tuning. Unlike previous approaches that incrementally improve SFT-trained models, \modelname enables RL training from scratch, allowing models to discover optimal tool utilization strategies through extensive exploration. This scaling approach yields qualitatively different behaviors than methods that build upon predetermined patterns.

Our experiments with Qwen2.5-Math base models at both 1.5B and 7B scales demonstrate substantial performance improvements over both traditional RL without code interpreters and Qwen2.5-Math-Instruct in TIR environments~\cite{yang2024qwen2}. Notably, our 7B model achieves a 43.3\% accuracy on AIME problems, comparable to some 32B models trained with reinforcement learning~\cite{OpenReasonerZero2025}.

Through extensive analysis, we uncover several key insights about tool-integrated learning dynamics:

\begin{itemize} 
\item \textbf{Code usage evolution:} As training progresses, the proportion of problems the model solves using code increases steadily, accompanied by a growing percentage of syntactically correct and executable code. This demonstrates the model's autonomous acquisition of effective tool utilization strategies.
\item \textbf{Self-regulation of ineffective code:} Without explicit instruction, models learn to identify and reduce the generation of ineffective code patterns, suggesting an emerging form of metacognition about tool utility.
\item \textbf{Tool call frequency trade-offs:} Increasing the maximum allowed tool calls per problem significantly enhances performance but introduces severe computational overhead, revealing a critical efficiency-effectiveness trade-off in tool-integrated learning.
\end{itemize}

These findings reveal how \modelname fosters diverse cognitive behaviors, including obtaining feedback from code execution, cross-checking between computational and analytical approaches, and adaptively selecting reasoning strategies based on problem characteristics. The emergence of these behaviors through reward-driven learning, rather than imitation of human-designed patterns, suggests tool-integrated reinforcement learning represents a promising direction for enhancing LLMs' reasoning capabilities.

To support further research, we open-source our implementation and trained models, enabling the community to build upon and extend \modelname for advancing tool-augmented language models.

\section{Methodology}
\subsection{Dataset Construction}
We constructed our dataset from Olympic-level mathematical competitions questions sourced from NuminaMATH~\cite{li2024numinamath}, MATH~\cite{hendrycks2021measuring}, and DeepScaleR~\cite{deepscaler2025}. The initial filtering removed proof-based problems and questions with ambiguous verification criteria, yielding 75,149 verifiable questions. We then applied LIMR~\cite{li2025limr}, a reinforcement learning data distillation technique, to extract high-quality samples with balanced difficulty distribution. This process resulted in a final dataset of 28,740 questions, which serves as the foundation for all subsequent experiments.

\subsection{Tool Intergrated Reasoning}
\label{subsec:method tir}

\begin{figure}[t]
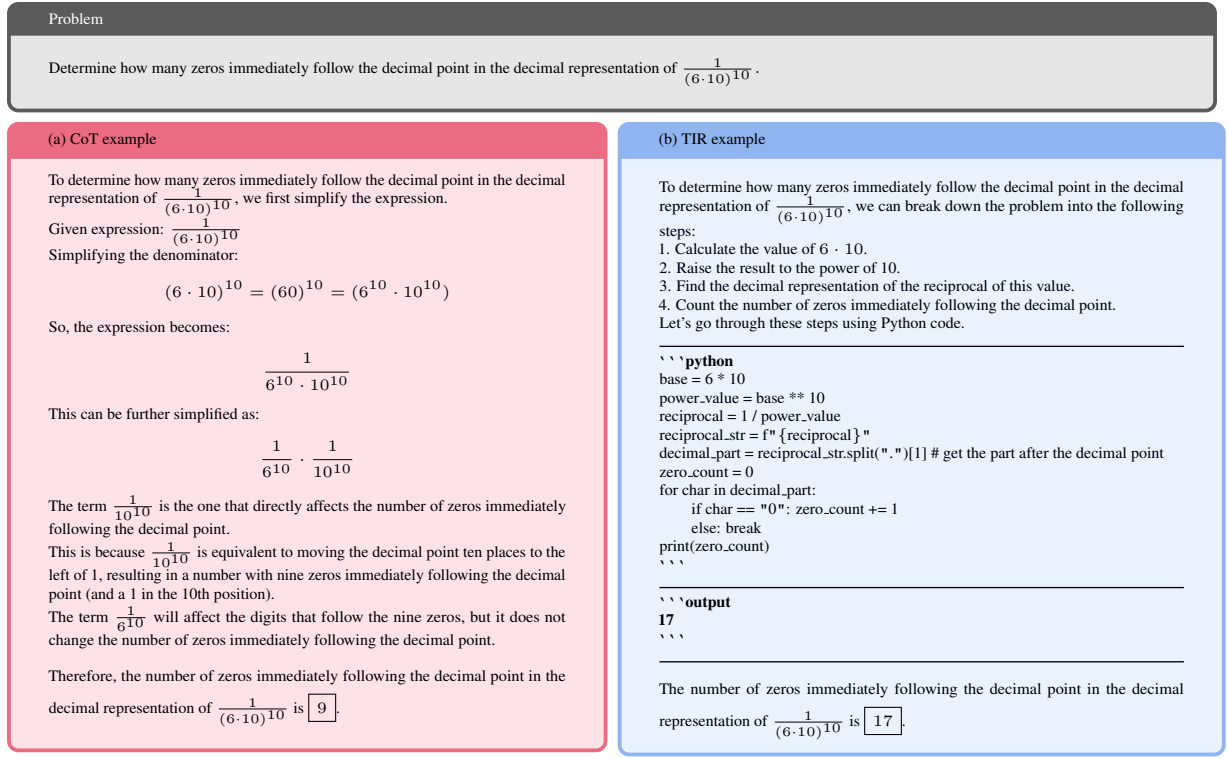

\setlength{\columnsep}{1.5mm}
\begin{tcolorbox}
[colback=grey!20!white,colframe=grey!70!black,title=\tiny{Problem}]
\begin{tiny}
Determine how many zeros immediately follow the decimal point in the decimal representation of $\frac{1}{(6 \cdot 10)^{10}}.$
\end{tiny}
\end{tcolorbox}
\vspace{-4.5mm}
\begin{multicols}{2}
\begin{tcolorbox}[colback=cot!20!white,colframe=cot!95!black,width=7.9cm,title=\tiny\textcolor{black}{(a) CoT example}, after=\vspace{4pt}]
\vspace{-1mm}
\begin{tiny}
To determine how many zeros immediately follow the decimal point in the decimal representation of \(\frac{1}{(6 \cdot 10)^{10}}\), we first simplify the expression.

Given expression:  \(\frac{1}{(6 \cdot 10)^{10}}\)

Simplifying the denominator:

\[(6 \cdot 10)^{10} = (60)^{10} = (6^{10} \cdot 10^{10})\]

So, the expression becomes:

\[\frac{1}{6^{10} \cdot 10^{10}}\]

This can be further simplified as:

\[\frac{1}{6^{10}} \cdot \frac{1}{10^{10}}\]

The term \(\frac{1}{10^{10}}\) is the one that directly affects the number of zeros immediately following the decimal point. 

\vspace{0.6mm}

This is because \(\frac{1}{10^{10}}\) is equivalent to moving the decimal point ten places to the left of 1, resulting in a number with nine zeros immediately following the decimal point (and a 1 in the 10th position).

\vspace{0.6mm}

The term \(\frac{1}{6^{10}}\) will affect the digits that follow the nine zeros, but it does not change the number of zeros immediately following the decimal point.

\vspace{0.6mm}

Therefore, the number of zeros immediately following the decimal point in the decimal representation of \(\frac{1}{(6 \cdot 10)^{10}}\) is \(\boxed{9}\).
\end{tiny}
\vspace{0mm}
\end{tcolorbox}

\begin{tcolorbox}[colback=tir!20!white,colframe=tir!95!black,width=8cm,title=\tiny\textcolor{black}{(b) TIR example}, after=\vspace{4pt}]
\begin{tiny}
To determine how many zeros immediately follow the decimal point in the decimal representation of \(\frac{1}{(6 \cdot 10)^{10}}\), we can break down the problem into the following steps:

1. Calculate the value of \(6 \cdot 10\).

2. Raise the result to the power of 10.

3. Find the decimal representation of the reciprocal of this value.

4. Count the number of zeros immediately following the decimal point.

Let's go through these steps using Python code.

\textcolor{black}{\rule{\linewidth}{0.5pt}}

\textbf{\texttt{```}python}

base = 6 * 10

power\_value = base ** 10

reciprocal = 1 / power\_value

reciprocal\_str = f\texttt{"}\{reciprocal\}\texttt{"}

decimal\_part = reciprocal\_str.split(\texttt{"}.\texttt{"})[1]  \# get the part after the decimal point

zero\_count = 0 

for char in decimal\_part:

    \hspace{2em}if char == \texttt{"}0\texttt{"}: zero\_count += 1
        
    \hspace{2em}else: break

print(zero\_count)

\textbf{\texttt{```}}

\textcolor{black}{\rule{\linewidth}{0.5pt}}

\textbf{\texttt{```}output}

\textbf{17}

\textbf{\texttt{```}}

\textcolor{black}{\rule{\linewidth}{0.5pt}}

\textcolor{black}{The number of zeros immediately following the decimal point in the decimal representation of \(\frac{1}{(6 \cdot 10)^{10}}\) is \(\boxed{17}\).}
\end{tiny}
\vspace{-1.2mm}
\end{tcolorbox}

\end{multicols}
\vspace{-4mm}
\caption{Example of CoT and TIR solution of the problem. TIR enables the model to write code and call an interpreter to obtain the output of the executed code, and then perform further reasoning based on the execution results.}

\label{fig:examples1}
\end{figure}




Tool Integrated Reasoning (TIR)~\cite{gou2023tora, wang2023mathcoder, shao2024deepseekmath, liao2024mario, yang2024qwen2} enables large language models (LLMs) to overcome their computational limitations by incorporating executable code into their reasoning process. Unlike traditional prompting approaches, TIR significantly improves accuracy on numerical computation and logical tasks through an iterative process of reasoning and code execution.

Formally, given a language model $M$, A code interpreter $I$ which executes code written by $M$ to obtain the results of tool calls, and an input question $Q$, TIR constructs a reasoning trajectory $s_k$ at step $k$ as:
\begin{align}
s_k = {r_1, c_1, o_1, \dots, r_k, c_k, o_k}
\end{align}
where $r_i$ represents natural language reasoning, $c_i$ denotes generated code, and $o_i$ is the execution result of $c_i$. The iterative generation process follows:
\begin{align}
&(r_{k}, c_{k}) = M(Q \oplus s_{k-1})\\
& o_{k} = I(c_{k})\\
& s_{k} = s_{k-1} \oplus r_{k} \oplus c_{k} \oplus o_{k}
\end{align}

This cycle continues until the model produces a final answer, with each step informed by previous code execution results. The dynamic adjustment of reasoning paths based on computational feedback allows the model to verify intermediate steps and correct errors during complex problem-solving.

\subsection{\modelname: Tool Integrated Reinforcement Learning}
We propose Tool Integrated Reinforcement Learning (\modelname), a framework that combines TIR with reinforcement learning directly from base language models without prior supervised fine-tuning. While existing approaches typically apply RL on models already aligned through SFT, our method demonstrates that tool-augmented reasoning capabilities can emerge effectively from base models through properly designed reinforcement learning.

In conventional RL for reasoning, models generate Chain-of-Thought (CoT) trajectories and receive rewards based on answer correctness. However, CoT often fails in scenarios requiring precise computation or exhaustive enumeration. Our approach replaces these pure language trajectories with tool-integrated reasoning paths, where a Code Interpreter module is integrated directly into the RL environment interaction loop.

\subsubsection{TIR Rollout Framework}


To enable the model to automatically output reasoning with code blocks, we use the prompt in Figure ~\ref{fig:prompt template}. During the model's rollout process, when a code termination identifier (\texttt{```output}) is detected, the system pauses text generation, extracts the latest code block for execution, and inserts the structured execution result into the context in the format \texttt{```output\textbackslash nOBSERVATION\textbackslash n```\textbackslash n}, where \texttt{OBSERVATION} is the execution result. The system then continues to generate subsequent natural language reasoning until the model either provides a final answer or produces a new code block. Notably, we deliberately return error messages to the LLM when code execution fails, as we hypothesize that these error diagnostics enhance the model's capacity to generate syntactically and semantically correct code in subsequent iterations.

\begin{figure}[htbp]
\centering
\begin{tcolorbox}[title=Prompt, width=16cm, boxrule=0.45mm, fonttitle=\bfseries]
A conversation between User and Assistant. The user asks a question, and the Assistant solves it.\textbackslash nUser: Please integrate natural language reasoning with programs to solve the problem above, and put your final answer within \textbackslash boxed\{\}.\textbackslash n\textcolor{red}{prompt}\textbackslash nAssistant:
\end{tcolorbox}
\caption{Prompt template for \modelname, and \textcolor{red}{prompt} will be replaced with specific question in training.}
\label{fig:prompt template}
\end{figure}

We observed that the integration of tool during the rollout process introduces significant GPU idle time, with speed of rollout process inversely proportional to the frequency of tool calls. To maintain training speed within a reasonable range, we introduced a hyperparameter, $C$, which is the maximum number of tool calls the model can make during a single response generation. Once this threshold is exceeded, the system ignores further code execution requests, forcing the model to switch to pure-text reasoning mode.

\subsubsection{Design Choices of \modelname}
\vspace{0.1cm}
\begin{enumerate}
    \item \textbf{Tool Call Frequency Control.} We observed that tool integration during rollout introduces significant GPU idle time, with rollout speed inversely proportional to tool call frequency. To maintain reasonable training efficiency, we introduced a hyperparameter $C$, representing the maximum number of tool calls allowed per response generation. Once this threshold is exceeded, the system ignores further code execution requests, forcing the model to switch to pure-text reasoning mode.
    \item \textbf{Execution Environment Selection.} To balance training efficiency and effectiveness, we sought a stable, accurate, and responsive code interpreter implementation. We initially tested the Python executor from qwen-agent\footnote{https://github.com/QwenLM/Qwen-Agent}, which offers extremely low latency. However, its execution environment lacks isolation from the training system, potentially allowing execution errors (e.g., segmentation faults from illegal memory access) to compromise the entire training process. We ultimately selected Sandbox Fusion\footnote{https://bytedance.github.io/SandboxFusion/}, which provides an isolated execution environment. Despite slightly higher latency, it delivers superior stability for sustained training operations.
    \item \textbf{Error Message Processing.} We implemented specific error handling optimizations to enhance training effectiveness. When Sandbox Fusion encounters execution errors, it generates verbose tracebacks containing irrelevant file path information. To reduce context length and preserve only relevant error information, we extract only the final line of error messages (e.g., \texttt{NameError: name 'a' is not defined}).
    \item \textbf{Sandbox Output Masking.} During loss computation, we mask out the \texttt{OBSERVATION} outputs from the sandbox environment, significantly improving training stability by preventing the model from attempting to memorize specific execution outputs rather than learning generalizable reasoning patterns.
\end{enumerate}



\subsubsection{Reward Design}

We implemented a rule-based reward function where correct answers receive a reward of 1, incorrect answers receive -1.
In addition, the code interpreter naturally provides feedback on code executability. Based on the correlation between successful code execution and problem-solving accuracy, we introduced an \textbf{execution-based penalty}: responses containing non-executable code incur a -0.5 reward reduction.

\noindent
\begin{minipage}[t]{0.48\textwidth} 
\centering
\captionof{table}{Answer Correctness Reward}
\begin{tabular}{cc}
\toprule
Answer Correctness & Reward Value \\
\midrule
Correct & 1 \\
Incorrect & -1 \\
\bottomrule
\end{tabular}

\end{minipage}
\hfill 
\begin{minipage}[t]{0.48\textwidth} 
\centering
\captionof{table}{Code Executability Reward}
\begin{tabular}{cc}
\toprule
Code Executability & Reward Value \\
\midrule
Executable & 0 \\
Unexecutable & -0.5 \\
\bottomrule
\end{tabular}
\end{minipage}


\section{Experiment}
\subsection{Experimental Setup}


\paragraph{Training}
All RL experiments are conducted using the veRL~\cite{sheng2024hybridflow} framework, with Sandbox Fusion as the code interpreter. We employ the GRPO~\cite{shao2024deepseekmath} algorithm, setting the rollout batch size to 128 and generating 16 samples per problem. To enhance model exploration, all experiments omit the KL loss and set the temperature to 1.
We choose the Qwen-2.5-Math~\cite{yang2024qwen2} series models as the base models for RL.
To maximize efficiency, the default number of calls, $C$, is set to \textbf{1}. Moreover, only the \textbf{Answer Correctness Reward} is retained in the default experiments, without incorporating the \textbf{Code Executability Reward}, which will be discussed in \S~\ref{subsec:analysis}.

\paragraph{Evaluation}
For evaluation, we use greedy decoding (temperature = 0) across all models. We evaluate our results on several challenging mathematical benchmarks: (1) AIME24, (2) AIME25, (3) MATH500~\cite{hendrycks2021measuring}, (4) OlympiadBench~\cite{he2024olympiadbench}, and (5) AMC23.

\subsection{Main Results}
\label{subsec:main-res}

\definecolor{lightblue}{rgb}{0.878, 0.947, 0.952}
\definecolor{ForestGreen}{rgb}{0.133, 0.545, 0.133}
\newcommand{\greentick}{\textcolor{green}{\ding{51}}}
\newcommand{\redcross}{\textcolor{red}{\ding{55}}}
\begin{table*}[htbp]
\centering
\caption{Comparison of different models testing accuracy on mathematical benchmarks. Qwen2.5-Math-1.5B-Instruct-TIR represents the testing of Qwen2.5-Math-1.5B-Instruct in the TIR environment. For a fair comparison, we set the maximum number of tool calls to 1. The best results are presented in \textbf{bold}.}
\label{table: main-result}
\resizebox{1.0\textwidth}{!}{
    \begin{tabular}{lcccccccc}
        \toprule
        Model & SFT/RL & Tool & AIME24 & AIME25 & MATH500 & Olympiad & AMC23 & Avg \\ 
        \midrule
        \multicolumn{8}{c}{\textit{Models based on Qwen2.5-Math-1.5B-Base}} \\
        Qwen2.5-Math-1.5B-Instruct & RL & \redcross  & 10.0 & 10.0 & 66.0 & 31.0 & 62.5 & 35.9\\
        Qwen2.5-Math-1.5B-Instruct-TIR & RL & \greentick  & 13.3 & 13.3 & 73.8 & 41.3 & 55.0 & 41.3\\
        \rowcolor{tir!20}\modelname-1.5B(Ours) & RL & \greentick & \textbf{26.7}$_{\textcolor{ForestGreen}{+13.3}}$ & \textbf{26.7}$_{\textcolor{ForestGreen}{+13.3}}$ & \textbf{77.8}$_{\textcolor{ForestGreen}{+3.0}}$ & \textbf{44.0}$_{\textcolor{ForestGreen}{+2.7}}$ & \textbf{67.5}$_{\textcolor{ForestGreen}{+5.0}}$ & \textbf{48.5}$_{\textcolor{ForestGreen}{+7.2}}$ \\ 
        \midrule
        \multicolumn{8}{c}{\textit{Models based on Qwen2.5-Math-7B-Base}} \\
        Qwen2.5-Math-7B-Instruct & RL & \redcross  & 10.0 & 16.7 & 74.8 & 32.4 & 65.0 & 39.8\\
        Qwen2.5-Math-7B-Instruct-TIR & RL & \greentick  & 26.7 & 16.7 & 78.8 & 45.0 & 70.0 & 47.4\\
        SimpleRL-Zero & RL & \redcross & 33.3 & 6.7 & 77.2 & 37.6 & 62.5 & 43.5\\ 
        rStar-Math-7B & SFT & \redcross & 26.7 & - & 78.4 & 47.1 & 47.5 & -\\ 
        Eurus-2-7B-PRIME & RL & \redcross & 26.7 & 13.3 & 79.2 & 42.1 & 57.4 & 43.1\\ 
        \rowcolor{tir!20}\modelname-7B(Ours) & RL & \greentick & \textbf{43.3}$_{\textcolor{ForestGreen}{+10.0}}$ & \textbf{30.0}$_{\textcolor{ForestGreen}{+13.3}}$ & \textbf{82.2}$_{\textcolor{ForestGreen}{+3.0}}$ & \textbf{49.9}$_{\textcolor{ForestGreen}{+2.8}}$ & \textbf{75.0}$_{\textcolor{ForestGreen}{+5.0}}$ & \textbf{62.1}$_{\textcolor{ForestGreen}{+14.7}}$ \\ 
        \bottomrule
    \end{tabular}
}
\end{table*}

\begin{figure}[ht]
    \centering
    \includegraphics[width=0.98\textwidth]{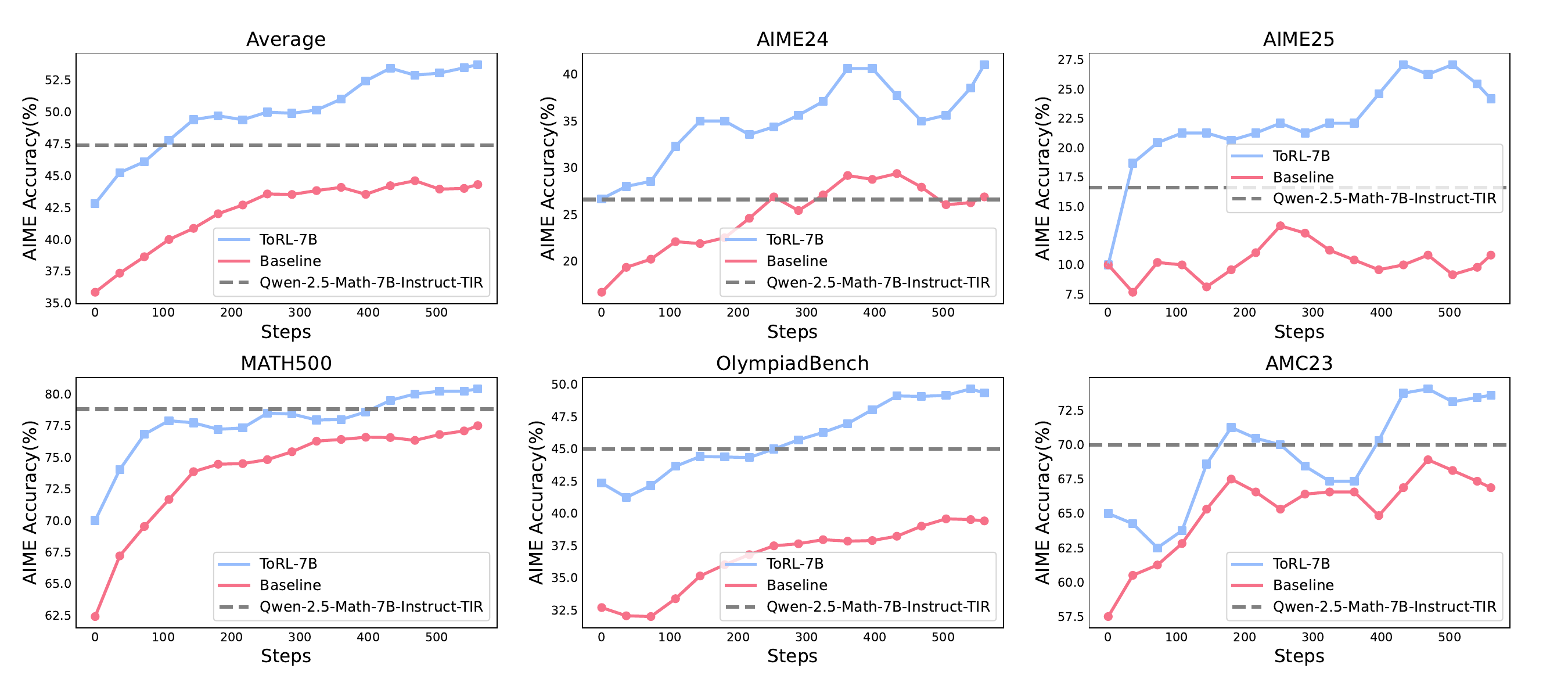} 
    \caption{Performance comparison across mathematical benchmarks. The plots show accuracy (\%) against training steps for the 7B model evaluated on different benchmarks. Across all benchmarks, \modelname-7B (blue) consistently outperforms both the baseline without tool integration (red) and Qwen-2.5-Math-7B-Instruct-TIR (dashed gray).}
    \label{fig:main-1-1}
\end{figure}

Table \ref{table: main-result} presents the performance comparison of different models on mathematical benchmarks. \modelname consistently outperforms baseline models across all tested benchmarks. For the 1.5B parameter models, \modelname-1.5B achieves an average accuracy of \textbf{48.5\%}, surpassing both Qwen2.5-Math-1.5B-Instruct (35.9\%) and Qwen2.5-Math-1.5B-Instruct-TIR (41.3\%). The improvement is even more pronounced in the 7B parameter models, where \modelname-7B reaches 62.1\% average accuracy, significantly higher than other open-source models with the same base model—representing an impressive \textbf{14.7\%} absolute improvement.

Figure \ref{fig:main-1-1} illustrates the training dynamics across five different mathematical benchmarks. \modelname-7B shows consistent improvement over training steps and maintains a clear advantage over both the baseline without tool integration and Qwen-2.5-Math-7B-Instruct-TIR. This performance gap is particularly notable in challenging benchmarks like AIME24, AIME25, and OlympiadBench, where \modelname-7B achieves \textbf{43.3\%}, \textbf{30.0\%}, and \textbf{49.9\%} accuracy respectively.

\subsection{Analysis}
\label{subsec:analysis}

To gain deeper insights into how RL facilitates the model's tool invocation for reasoning assistance, we quantitatively analyze the model's tool usage and code-writing behaviors in \S~\ref{subsubsec:p1} and showcase illustrative examples in \S~\ref{subsubsec:p3}. Furthermore, we change critical parameters related to \modelname in the RL framework to examine their effects on both model performance and training efficiency in \S~\ref{subsubsec:p2}.

\subsubsection{Part I: Code behaviors during training}
\label{subsubsec:p1}
\begin{figure}[htbp]
    \centering
    \begin{subfigure}[b]{0.245\textwidth}
        \includegraphics[width=\textwidth]{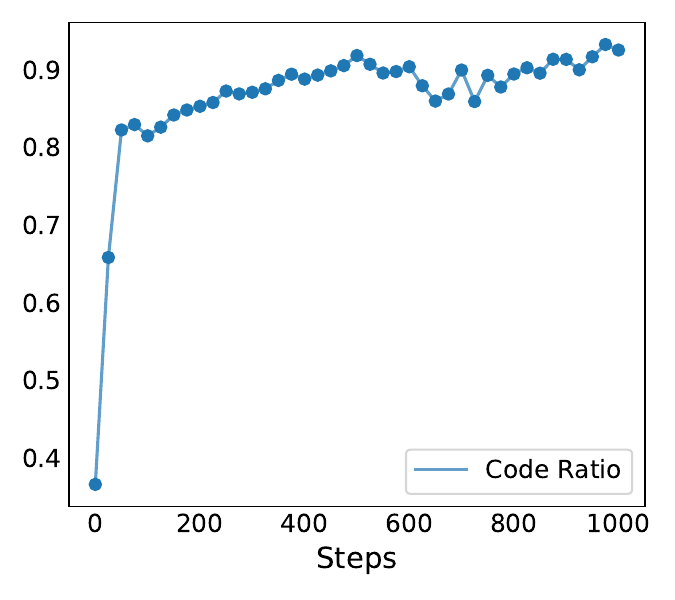}
        \caption{Code Ratio}
        \label{fig:analysis-1}
    \end{subfigure}
    \begin{subfigure}[b]{0.245\textwidth}
        \includegraphics[width=\textwidth]{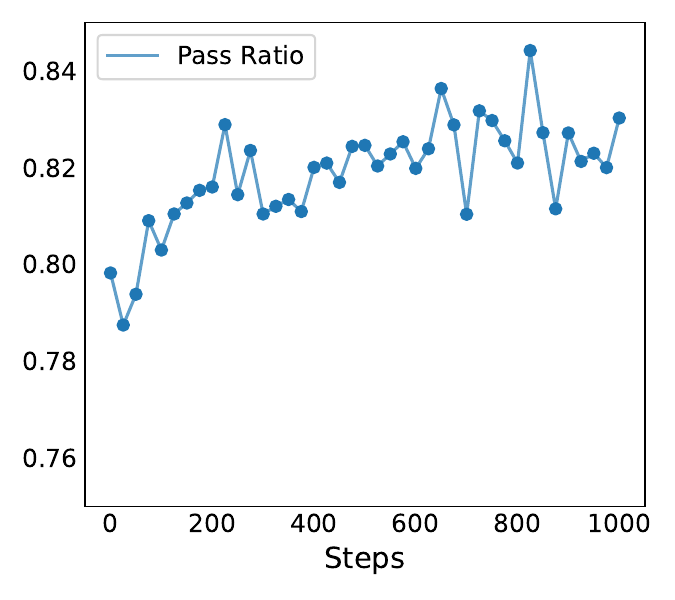}
        \caption{Pass Ratio}
        \label{fig:analysis-2}
    \end{subfigure}
    \begin{subfigure}[b]{0.245\textwidth}
        \includegraphics[width=\textwidth]{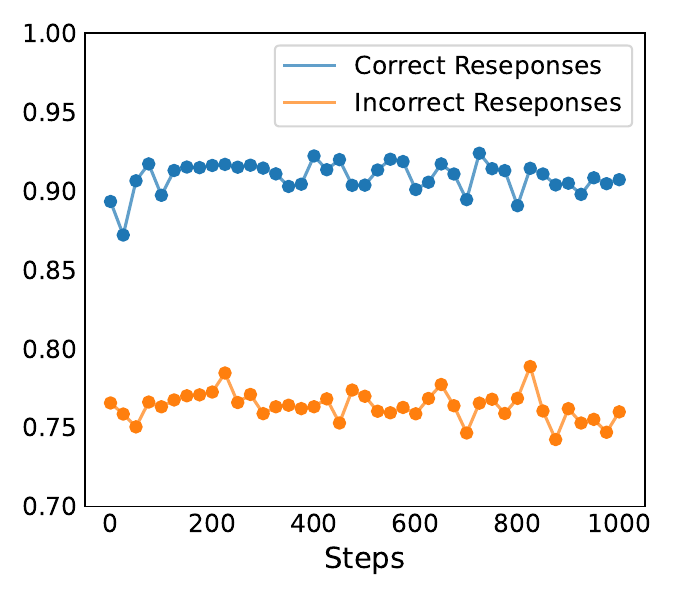}
        \caption{Correct/Incorrect Pass Ratio}
        \label{fig:analysis-3}
    \end{subfigure}
    \begin{subfigure}[b]{0.245\textwidth}
        \includegraphics[width=\textwidth]{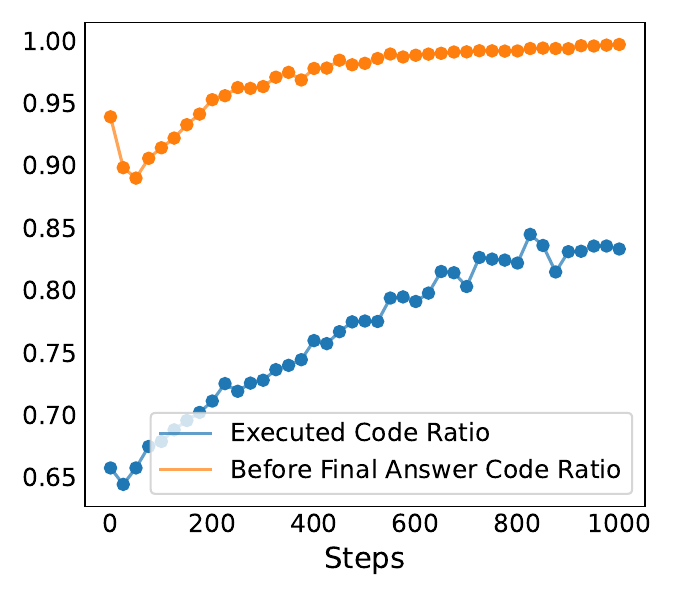}
        \caption{Effective Pass Ratio}
        \label{fig:analysis-4}
    \end{subfigure}
    \caption{Tool usage dynamics during training. (a) Code Ratio: proportion of responses that contain code, (b) Pass Ratio: proportion of executed code that runs without errors, (c) Pass Ratio in Correct/Incorrect Responses: comparing execution success rates for correct vs. incorrect responses, and (d) Effective Code Ratio.} 
    \label{fig:analysis}
\end{figure}

Figure \ref{fig:analysis} provides insights into tool usage patterns during training, including:
\paragraph{Code Ratio} Figure \ref{fig:analysis-1} illustrates the proportion of model-generated responses containing code within the first 100 steps, which increased from 40\% to 80\%, demonstrating steady improvement throughout training.
\paragraph{Pass Ratio} Figure \ref{fig:analysis-2} presents the proportion of successfully executed code, highlighting a continuous upward trend that reflects the model's enhanced coding capability.
\paragraph{Pass Ratio of Correct to Incorrect Responses} Meanwhile, Figure \ref{fig:analysis-3} depicts the Pass Ratio of Correct to Incorrect Responses, revealing a correlation between code execution errors and final answer accuracy, with correct responses exhibiting higher code pass rates.
\paragraph{Effective Code Ratio} Furthermore, Figure \ref{fig:analysis-4} examines changes in the proportion of effective code, focusing on two key metrics: 
\begin{itemize}
    \item Executed code ratio, which represents the proportion of generated code successfully executed by the interpreter, excluding any code omitted due to due to exceeding the maximum number of tool calls ($C$). 
    \item Before final answer code ratio, which measures the proportion of code generated before the model provides its final answer, disregarding code used solely for correctness verification post-answer generation.
\end{itemize}
Both metrics increased over time, indicating that the effectiveness of model-generated code improved as training progressed.

\begin{tcolorbox}[title=Takeaway-I, width=16cm, boxrule=0.45mm, fonttitle=\bfseries]
As the number of training steps increases, the proportion of problems the model solves using code, as well as the proportion of code that can be executed correctly, continues to grow. Meanwhile, during the training process, the model identifies and reduces the generation of ineffective code.
\end{tcolorbox}

\subsubsection{Part II: Impact of Key Settings}
\label{subsubsec:p2}
\begin{figure}[htbp]
    \centering
    \begin{subfigure}[b]{0.245\textwidth}
        \includegraphics[width=\textwidth]{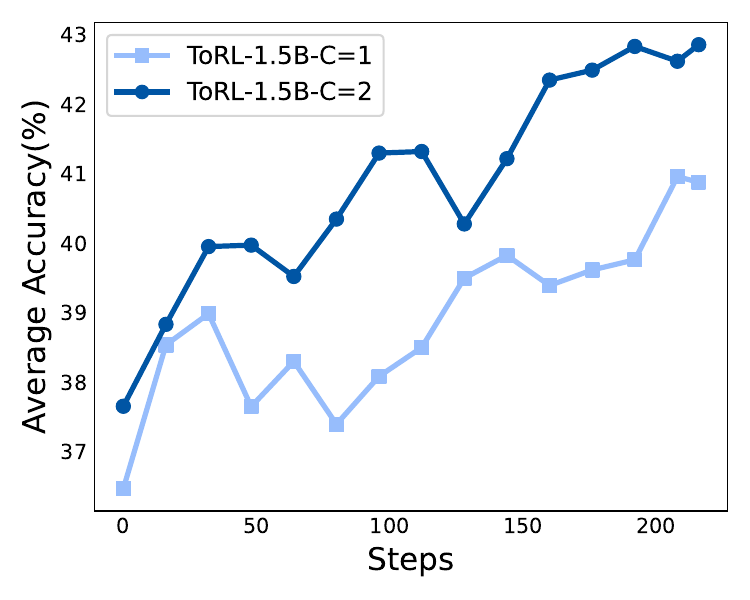}
        \caption{}
        \label{fig:ablation-1}
    \end{subfigure}
    \hfill
    \begin{subfigure}[b]{0.245\textwidth}
        \includegraphics[width=\textwidth]{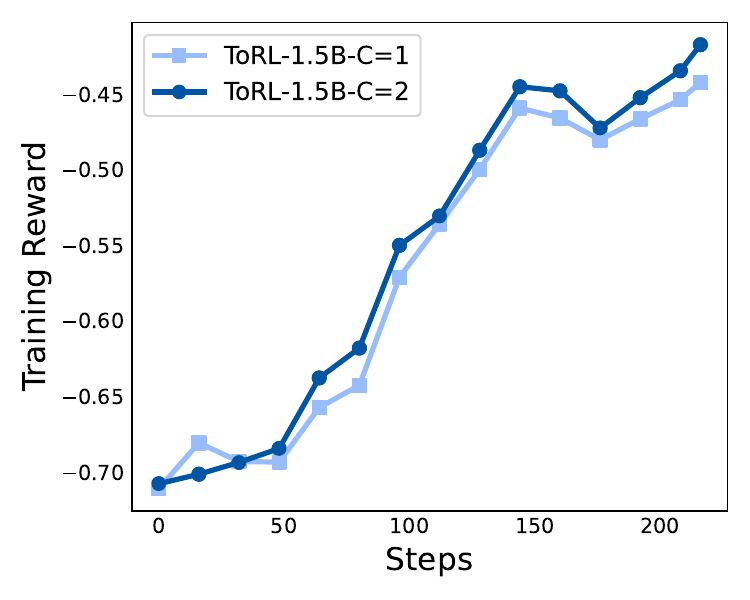}
        \caption{}
        \label{fig:ablation-2}
    \end{subfigure}
    \hfill
    \begin{subfigure}[b]{0.245\textwidth}
        \includegraphics[width=\textwidth]{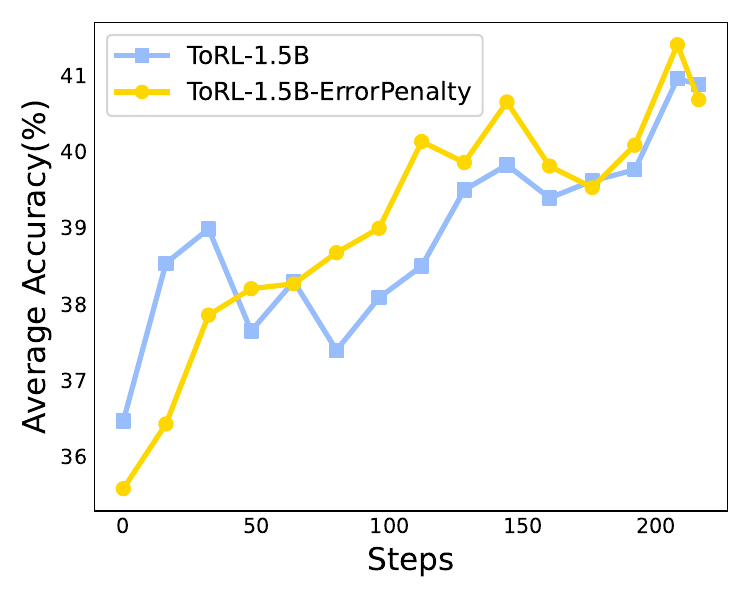}
        \caption{}
        \label{fig:ablation-3}
    \end{subfigure}
    \hfill
    \begin{subfigure}[b]{0.245\textwidth}
        \includegraphics[width=\textwidth]{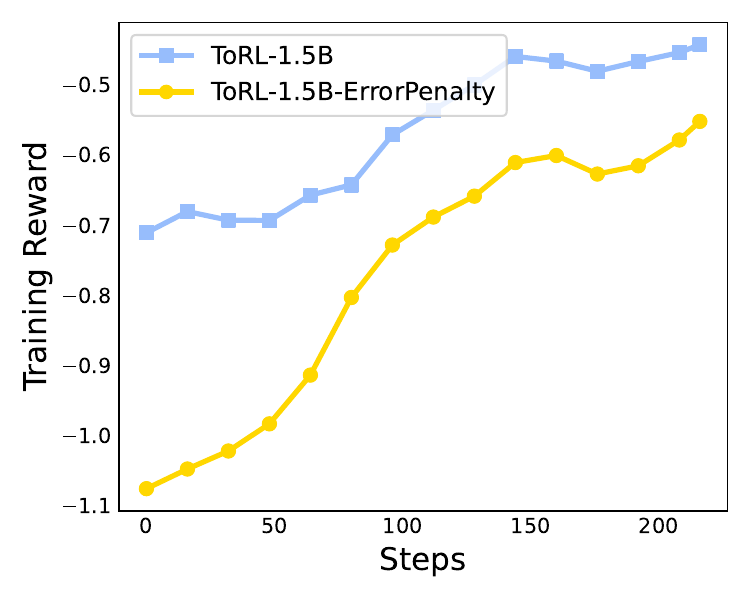}
        \caption{}
        \label{fig:ablation-4}
    \end{subfigure}
    \caption{\modelname-1.5B-C=2 indicates that the hyperparameter \( C \) is set to 2. \modelname-1.5B-ErrorPenalty represents the incorporation of the Code Executability Reward into the reward function, penalizing incorrectly executed code.} 
    \label{fig:ablation}
\end{figure}


In this section, we explore the impact of key \modelname settings on final performance and behavior.

\begin{wraptable}{r}{0.3\textwidth}
\centering
\small
\begin{tabular}{cr}
\toprule
\textbf{$C$} & \textbf{Average Step Time(s)} \\ 
\midrule
0  & 118 \\
1  & 237 \\ 
2  & 288 \\ 
\bottomrule
\end{tabular}
\caption{The average time per single step under different values of $C$(measured on an 8*A800 GPU system). }
\label{tab: step time}
\end{wraptable}

First, we investigate the effect of increasing $C$, the maximum number of tools that can be called in a single response generation during rollout. Specifically, we examine the impact of setting $C$ to 1 and 2 on model performance. As shown in Figures ~\ref{fig:ablation-1} and ~\ref{fig:ablation-2}, setting $C$ to 2 significantly improves performance, increasing average accuracy by approximately 2\%. However, as indicated in Table ~\ref{tab: step time}, increasing $C$ substantially reduces training speed, necessitating a trade-off between performance and efficiency.

Furthermore, we analyze the impact of incorporating \textbf{Code Executability Reward} into the reward shaping. Figures ~\ref{fig:ablation-3} and ~\ref{fig:ablation-4} demonstrate that this reward design does not enhance model performance. We hypothesize that introducing penalties for execution errors may incentivize the model to generate overly simplistic code to minimize errors, which could, in turn, hinder its ability to solve problems correctly.

\begin{tcolorbox}[title=Takeaway-II, width=16cm, boxrule=0.45mm, fonttitle=\bfseries] 
Increasing the $C$ can significantly enhance model performance but also severely reduces training efficiency. Incorporating \textbf{Code Executability Reward} does not improve performance. 
\end{tcolorbox}

\subsubsection{Part III: Cognitive Behavior via RL Scaling}
\label{subsubsec:p3}
\begin{table}[t]
\tiny
    \centering
    \begin{minipage}{0.48\textwidth}
        \centering
        \caption{The code initially generate by the model encountered an execution error, then it is corrected by model and is successfully executed.}
        
        \begin{tabular}{l}
    
\toprule 
Question: Given the polynomial $f(x)=x^{6}-5x^{5}+6x^{4}+x^{2}+0.3x+2$, \\
use Horner's method to calculate $f(-2)$ and find the value of $v_{1}$. \\
\midrule
Response: To solve the given problem using Horner's method, we can \ldots \\
\texttt{```python} \\
\texttt{def horner(coeffs, x):} \\
    \ldots \\
\texttt{    return result, \textcolor{red}{results[1]}} \\
\texttt{```} \\
\texttt{```output} \\
\texttt{TypeError: 'int' object is not subscriptable} \\
\texttt{```} \\
To fix this, we need to ensure that we properly store and access the \ldots \\
\texttt{```python} \\
\texttt{def horner(coeffs, x):} \\
\texttt{    }\ldots \\
\texttt{    return result, \textcolor{red}{results}} \\
\texttt{```} \\
\texttt{```output} \\
\texttt{v\_1 = -7} \\
\texttt{```} \\
Thus, the value of \( f(-2) \) is \(\boxed{325.4}\) and the value of \( v_1 \) is \(\boxed{-7}\). \\
\bottomrule

\end{tabular}

        \label{tab:case1}
    \end{minipage}%
    \hfill
    \begin{minipage}{0.48\textwidth}
    \caption{The model first derives an incorrect result based on natural language reasoning, then discovers the error during code verification and makes corrections}
        \centering
        \begin{tabular}{l}
            \toprule 
Question: After labeling 12 balls with numbers 1, 2, 3, ... 12 \ldots \\
 What are the numbers on the other three balls taken by Person C? \\
\midrule
Response: To solve the problem, we need to find the numbers on the  \ldots \\
\ldots  \\
The numbers on the other three balls taken by Person C are \(2, 10, 13\), \\ 
reduced to valid numbers from 2 to 12, so the only valid combination is \textcolor{red}{\(2, 10, 13\)}. \\
\ldots \\
Here is the Python code to verify the solution: \\
\texttt{```python} \\
\texttt{\# Define the total sum of the numbers from 1 to 12} \\
\texttt{total\_sum = sum(range(1, 13))} \\
\texttt{\# Each person's sum is total\_sum / 3} \\
\texttt{person\_sum = total\_sum // 3} \\
\ldots \\
\texttt{\# Output the balls taken by Person C} \\
\texttt{print(balls\_C\_remaining)} \\
\texttt{```output} \\
\textcolor{red}{\texttt{[3, 10, 12]}} \\
\texttt{```} \\
\ldots \\
The numbers on the other three balls taken by Person C are \textcolor{red}{\(\boxed{3, 10, 12}\).} \\
\bottomrule

        \end{tabular}
        \label{tab:case2}
    \end{minipage}
\end{table}


We present some interesting phenomena that occurred in the later stages of model training, which can help us further understand in detail the model's cognitive behavior in solving problems using tools.

Table~\ref{tab:case1} shows an example of the model adjusting its reasoning based on execution feedback from the code interpreter. The model first wrote the code, but improper handling of \texttt{results} caused an indexing error. After receiving the feedback \texttt{TypeError: 'int' object is not subscriptable}, it promptly adjusted and produced executable code, eventually inferring the correct answer.

Table~\ref{tab:case2} demonstrates the model’s cognitive behavior of reflection. The model initially solved the problem with natural language reasoning, then verified it through tool but found inconsistencies. Consequently, the model further makes corrections, ultimately generate the correct answer.

\begin{tcolorbox}[title=Takeaway-III, width=16cm, boxrule=0.45mm, fonttitle=\bfseries] 
\modelname gives rise to a variety of cognitive behaviors, including obtaining feedback from code execution results, cross-checking with code and natural language.
\end{tcolorbox}

\section{Conclusion}
We propose \modelname, which enables LLMs to integrate tools in reasoning through reinforcement learning, surpassing predefined tool usage constraints. Our results show significant performance gains and emergent reasoning, highlighting \modelname's potential for advancing LLMs in complex reasoning.

\newpage

\bibliographystyle{acl_natbib}
\bibliography{related}

\begin{thebibliography}{22}
\expandafter\ifx\csname natexlab\endcsname\relax\def\natexlab#1{#1}\fi

\bibitem[{Chen et~al.(2022)Chen, Ma, Wang, and Cohen}]{chen2022program}
Wenhu Chen, Xueguang Ma, Xinyi Wang, and William~W Cohen. 2022.
\newblock Program of thoughts prompting: Disentangling computation from reasoning for numerical reasoning tasks.
\newblock \emph{arXiv preprint arXiv:2211.12588}.

\bibitem[{Gao et~al.(2023)Gao, Madaan, Zhou, Alon, Liu, Yang, Callan, and Neubig}]{gao2023pal}
Luyu Gao, Aman Madaan, Shuyan Zhou, Uri Alon, Pengfei Liu, Yiming Yang, Jamie Callan, and Graham Neubig. 2023.
\newblock Pal: Program-aided language models.
\newblock In \emph{International Conference on Machine Learning}, pages 10764--10799. PMLR.

\bibitem[{Gou et~al.(2023)Gou, Shao, Gong, Shen, Yang, Huang, Duan, and Chen}]{gou2023tora}
Zhibin Gou, Zhihong Shao, Yeyun Gong, Yelong Shen, Yujiu Yang, Minlie Huang, Nan Duan, and Weizhu Chen. 2023.
\newblock Tora: A tool-integrated reasoning agent for mathematical problem solving.
\newblock \emph{arXiv preprint arXiv:2309.17452}.

\bibitem[{Guo et~al.(2025)Guo, Yang, Zhang, Song, Zhang, Xu, Zhu, Ma, Wang, Bi et~al.}]{guo2025deepseek}
Daya Guo, Dejian Yang, Haowei Zhang, Junxiao Song, Ruoyu Zhang, Runxin Xu, Qihao Zhu, Shirong Ma, Peiyi Wang, Xiao Bi, et~al. 2025.
\newblock Deepseek-r1: Incentivizing reasoning capability in llms via reinforcement learning.
\newblock \emph{arXiv preprint arXiv:2501.12948}.

\bibitem[{He et~al.(2024)He, Luo, Bai, Hu, Thai, Shen, Hu, Han, Huang, Zhang et~al.}]{he2024olympiadbench}
Chaoqun He, Renjie Luo, Yuzhuo Bai, Shengding Hu, Zhen~Leng Thai, Junhao Shen, Jinyi Hu, Xu~Han, Yujie Huang, Yuxiang Zhang, et~al. 2024.
\newblock Olympiadbench: A challenging benchmark for promoting agi with olympiad-level bilingual multimodal scientific problems.
\newblock \emph{arXiv preprint arXiv:2402.14008}.

\bibitem[{Hendrycks et~al.(2021)Hendrycks, Burns, Kadavath, Arora, Basart, Tang, Song, and Steinhardt}]{hendrycks2021measuring}
Dan Hendrycks, Collin Burns, Saurav Kadavath, Akul Arora, Steven Basart, Eric Tang, Dawn Song, and Jacob Steinhardt. 2021.
\newblock Measuring mathematical problem solving with the math dataset.
\newblock \emph{arXiv preprint arXiv:2103.03874}.

\bibitem[{Hu et~al.(2025)Hu, Zhang, Han, Jiang, and Xiangyu~Zhang}]{OpenReasonerZero2025}
Jingcheng Hu, Yinmin Zhang, Qi~Han, Daxin Jiang, and Heung-Yeung~Shum Xiangyu~Zhang. 2025.
\newblock Open-reasoner-zero: An open source approach to scaling reinforcement learning on the base model.
\newblock \url{https://github.com/Open-Reasoner-Zero/Open-Reasoner-Zero}.

\bibitem[{Huang et~al.(2024)Huang, Zou, Li, Liu, Zheng, Chern, Xia, Qin, Yuan, and Liu}]{huang2024o1}
Zhen Huang, Haoyang Zou, Xuefeng Li, Yixiu Liu, Yuxiang Zheng, Ethan Chern, Shijie Xia, Yiwei Qin, Weizhe Yuan, and Pengfei Liu. 2024.
\newblock O1 replication journey--part 2: Surpassing o1-preview through simple distillation, big progress or bitter lesson?
\newblock \emph{arXiv preprint arXiv:2411.16489}.

\bibitem[{Li et~al.(2024)Li, Beeching, Tunstall, Lipkin, Soletskyi, Huang, Rasul, Yu, Jiang, Shen et~al.}]{li2024numinamath}
Jia Li, Edward Beeching, Lewis Tunstall, Ben Lipkin, Roman Soletskyi, Shengyi Huang, Kashif Rasul, Longhui Yu, Albert~Q Jiang, Ziju Shen, et~al. 2024.
\newblock Numinamath: The largest public dataset in ai4maths with 860k pairs of competition math problems and solutions.
\newblock \emph{Hugging Face repository}, 13:9.

\bibitem[{Li et~al.(2025)Li, Zou, and Liu}]{li2025limr}
Xuefeng Li, Haoyang Zou, and Pengfei Liu. 2025.
\newblock Limr: Less is more for rl scaling.
\newblock \emph{arXiv preprint arXiv:2502.11886}.

\bibitem[{Liao et~al.(2024)Liao, Luo, Li, Wu, and Fan}]{liao2024mario}
Minpeng Liao, Wei Luo, Chengxi Li, Jing Wu, and Kai Fan. 2024.
\newblock Mario: Math reasoning with code interpreter output--a reproducible pipeline.
\newblock \emph{arXiv preprint arXiv:2401.08190}.

\bibitem[{Luo et~al.(2025)Luo, Tan, Wong, Shi, Tang, Roongta, Cai, Luo, Zhang, Li, Popa, and Stoica}]{deepscaler2025}
Michael Luo, Sijun Tan, Justin Wong, Xiaoxiang Shi, William~Y. Tang, Manan Roongta, Colin Cai, Jeffrey Luo, Tianjun Zhang, Li~Erran Li, Raluca~Ada Popa, and Ion Stoica. 2025.
\newblock Deepscaler: Surpassing o1-preview with a 1.5b model by scaling rl.
\newblock \url{https://pretty-radio-b75.notion.site/DeepScaleR-Surpassing-O1-Preview-with-a-1-5B-Model-by-Scaling-RL-19681902c1468005bed8ca303013a4e2}.
\newblock Notion Blog.

\bibitem[{Qin et~al.(2024)Qin, Li, Zou, Liu, Xia, Huang, Ye, Yuan, Liu, Li et~al.}]{qin2024o1}
Yiwei Qin, Xuefeng Li, Haoyang Zou, Yixiu Liu, Shijie Xia, Zhen Huang, Yixin Ye, Weizhe Yuan, Hector Liu, Yuanzhi Li, et~al. 2024.
\newblock O1 replication journey: A strategic progress report--part 1.
\newblock \emph{arXiv preprint arXiv:2410.18982}.

\bibitem[{Shao et~al.(2024)Shao, Wang, Zhu, Xu, Song, Bi, Zhang, Zhang, Li, Wu et~al.}]{shao2024deepseekmath}
Zhihong Shao, Peiyi Wang, Qihao Zhu, Runxin Xu, Junxiao Song, Xiao Bi, Haowei Zhang, Mingchuan Zhang, YK~Li, Y~Wu, et~al. 2024.
\newblock Deepseekmath: Pushing the limits of mathematical reasoning in open language models.
\newblock \emph{arXiv preprint arXiv:2402.03300}.

\bibitem[{Sheng et~al.(2024)Sheng, Zhang, Ye, Wu, Zhang, Zhang, Peng, Lin, and Wu}]{sheng2024hybridflow}
Guangming Sheng, Chi Zhang, Zilingfeng Ye, Xibin Wu, Wang Zhang, Ru~Zhang, Yanghua Peng, Haibin Lin, and Chuan Wu. 2024.
\newblock Hybridflow: A flexible and efficient rlhf framework.
\newblock \emph{arXiv preprint arXiv: 2409.19256}.

\bibitem[{Team et~al.(2025)Team, Du, Gao, Xing, Jiang, Chen, Li, Xiao, Du, Liao, Tang, Wang, Zhang, Yuan, Lu, Tang, Sung, Wei, Lai, Guo, Zhu, Ding, Hu, Yang, Zhang, Yao, Zhao, Lu, Li, Yu, Gao, Zheng, Yuan, Chen, Guo, Su, Wang, Zhao, Zhang, Liu, Yan, Wu, Shi, Ye, Yu, Dong, Zhang, Ma, Pan, Gong, Liu, Ma, Wei, Cao, Huang, Jiang, Gao, Xiong, He, Huang, Wu, He, Wei, Jia, Wu, Xu, Zu, Zhou, Pan, Charles, Li, Hu, Liu, Chen, Wang, Liu, Qin, Liu, Yang, Bao, Du, Wu, Wang, Zhou, Wang, Li, Zhu, Zhang, Wang, Yang, Huang, Huang, Xu, and Yang}]{kimiteam2025kimik15scalingreinforcement}
Kimi Team, Angang Du, Bofei Gao, Bowei Xing, Changjiu Jiang, Cheng Chen, Cheng Li, Chenjun Xiao, Chenzhuang Du, Chonghua Liao, Chuning Tang, Congcong Wang, Dehao Zhang, Enming Yuan, Enzhe Lu, Fengxiang Tang, Flood Sung, Guangda Wei, Guokun Lai, Haiqing Guo, Han Zhu, Hao Ding, Hao Hu, Hao Yang, Hao Zhang, Haotian Yao, Haotian Zhao, Haoyu Lu, Haoze Li, Haozhen Yu, Hongcheng Gao, Huabin Zheng, Huan Yuan, Jia Chen, Jianhang Guo, Jianlin Su, Jianzhou Wang, Jie Zhao, Jin Zhang, Jingyuan Liu, Junjie Yan, Junyan Wu, Lidong Shi, Ling Ye, Longhui Yu, Mengnan Dong, Neo Zhang, Ningchen Ma, Qiwei Pan, Qucheng Gong, Shaowei Liu, Shengling Ma, Shupeng Wei, Sihan Cao, Siying Huang, Tao Jiang, Weihao Gao, Weimin Xiong, Weiran He, Weixiao Huang, Wenhao Wu, Wenyang He, Xianghui Wei, Xianqing Jia, Xingzhe Wu, Xinran Xu, Xinxing Zu, Xinyu Zhou, Xuehai Pan, Y.~Charles, Yang Li, Yangyang Hu, Yangyang Liu, Yanru Chen, Yejie Wang, Yibo Liu, Yidao Qin, Yifeng Liu, Ying Yang, Yiping Bao, Yulun Du, Yuxin Wu, Yuzhi Wang, Zaida Zhou,
  Zhaoji Wang, Zhaowei Li, Zhen Zhu, Zheng Zhang, Zhexu Wang, Zhilin Yang, Zhiqi Huang, Zihao Huang, Ziyao Xu, and Zonghan Yang. 2025.
\newblock \href {http://arxiv.org/abs/2501.12599} {Kimi k1.5: Scaling reinforcement learning with llms}.

\bibitem[{Team(2024)}]{Openo1}
OpenO1 Team. 2024.
\newblock \href {https://github.com/Open-Source-O1/Open-O1} {Openo1}.
\newblock \emph{Github}.

\bibitem[{Wang et~al.(2023)Wang, Ren, Zhou, Lu, Luo, Shi, Zhang, Song, Zhan, and Li}]{wang2023mathcoder}
Ke~Wang, Houxing Ren, Aojun Zhou, Zimu Lu, Sichun Luo, Weikang Shi, Renrui Zhang, Linqi Song, Mingjie Zhan, and Hongsheng Li. 2023.
\newblock Mathcoder: Seamless code integration in llms for enhanced mathematical reasoning.
\newblock \emph{arXiv preprint arXiv:2310.03731}.

\bibitem[{Wei et~al.(2022)Wei, Wang, Schuurmans, Bosma, Xia, Chi, Le, Zhou et~al.}]{wei2022chain}
Jason Wei, Xuezhi Wang, Dale Schuurmans, Maarten Bosma, Fei Xia, Ed~Chi, Quoc~V Le, Denny Zhou, et~al. 2022.
\newblock Chain-of-thought prompting elicits reasoning in large language models.
\newblock \emph{Advances in neural information processing systems}, 35:24824--24837.

\bibitem[{Yang et~al.(2024)Yang, Zhang, Hui, Gao, Yu, Li, Liu, Tu, Zhou, Lin et~al.}]{yang2024qwen2}
An~Yang, Beichen Zhang, Binyuan Hui, Bofei Gao, Bowen Yu, Chengpeng Li, Dayiheng Liu, Jianhong Tu, Jingren Zhou, Junyang Lin, et~al. 2024.
\newblock Qwen2. 5-math technical report: Toward mathematical expert model via self-improvement.
\newblock \emph{arXiv preprint arXiv:2409.12122}.

\bibitem[{Ye et~al.(2025)Ye, Huang, Xiao, Chern, Xia, and Liu}]{ye2025limo}
Yixin Ye, Zhen Huang, Yang Xiao, Ethan Chern, Shijie Xia, and Pengfei Liu. 2025.
\newblock Limo: Less is more for reasoning.
\newblock \emph{arXiv preprint arXiv:2502.03387}.

\bibitem[{Zeng et~al.(2025)Zeng, Huang, Liu, He, Liu, Ma, and He}]{zeng2025simplerl}
Weihao Zeng, Yuzhen Huang, Wei Liu, Keqing He, Qian Liu, Zejun Ma, and Junxian He. 2025.
\newblock 7b model and 8k examples: Emerging reasoning with reinforcement learning is both effective and efficient.
\newblock \url{https://hkust-nlp.notion.site/simplerl-reason}.
\newblock Notion Blog.

\end{thebibliography}

\end{document}